\documentclass[10pt,twocolumn,letterpaper]{article}

\usepackage{cvpr}
\usepackage{times}
\usepackage{epsfig}
\usepackage{graphicx}
\usepackage{amsmath}
\usepackage{amssymb}

\usepackage{siunitx} 
\usepackage{multirow}
\usepackage{lipsum}
\usepackage{booktabs} 
\usepackage{tabularx}

\graphicspath{ {images/} }

\usepackage[breaklinks=true,bookmarks=false]{hyperref}

\cvprfinalcopy 

\def\cvprPaperID{****} 

\begin{document}

\title{\ BlockCNN: A Deep Network for Artifact Removal and Image Compression}
\author{Danial Maleki, Soheila Nadalian, Mohammad Mahdi Derakhshani, Mohammad Amin Sadeghi\\
School of Electrical and Computer Engineering, University of Tehran\\
{\tt\small \{d.maleki, nadalian.soheila, mderakhshani, asadeghi\}@ut.ac.ir}
}

\maketitle

\maketitle

\begin{abstract}
We present a general technique that performs both artifact removal and image compression. For artifact removal, we input a JPEG image and try to remove its compression artifacts. For compression, we input an image and process its $8 \times 8$ blocks in a sequence. For each block, we first try to predict its intensities based on previous blocks; then, we store a residual with respect to the input image. Our technique reuses JPEG's legacy compression and decompression routines. Both our artifact removal and our image compression techniques use the same deep network, but with different training weights. Our technique is simple and fast and it significantly improves the performance of artifact removal and image compression. 

\end{abstract}

\section{Introduction}
\label{sec:intro}

The advent of Deep Learning has led to multiple breakthroughs in image representation including: super-resolution, image compression, image enhancement and image generation. We present a unified model that can perform two tasks: 1- artifact removal for JPEG images and 2- image compression for new images. 

\begin{figure}[t]
\centering
    \includegraphics[width=8.2 cm]{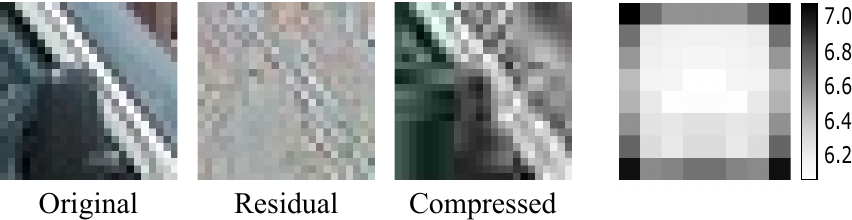}
    \caption{
    Left: JPEG compresses each $8 \times 8$ block independently. Therefore, each block has independent artifact characteristics. Our artifact removal technique acts on each block separately. Right: The statistics of a pixel's compression artifact depends on where it is located within an $8 \times 8$ block. The right figure, illustrates Mean Square Error of pixel intensities (within a block) after compression. We used 6 Million image blocks with quality factor of $20$ to produce this figure. }
\label{fig:motivation}
\end{figure}

Our model uses deep learning and legacy JPEG compression routines. JPEG divides images into $8 \times 8$ blocks and compresses each block independently. This causes block-wise compression artifact (Figure~\ref{fig:motivation}). We show that the statistics of a pixel's artifact depends on where it is placed in the block (Figure~\ref{fig:motivation}). As a result, an artifact removal technique that has a prior about the pixel's location has an advantage. Our model acts on $8 \times 8$ blocks in order to gain from this prior. Also, this let us reuse JPEG compression.

\begin{figure}[t]
\centering
    \includegraphics[width=7.5cm]{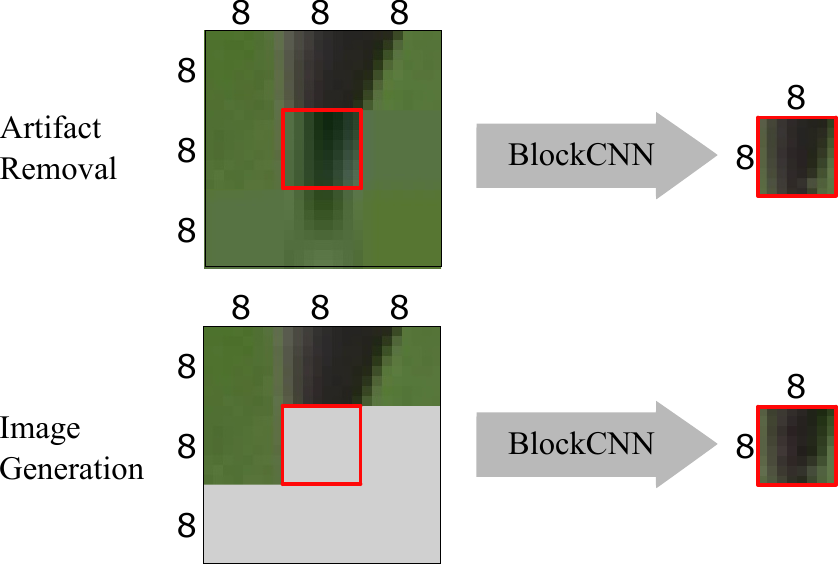}
    \caption{
    BlockCNN: This architecture can be used for both artifact removal and image compression. BlockCNN acts on $8 \times 8$ image blocks. Top: To remove artifact from each block, we input this block together with its eight adjacent blocks and try to removes artifacts from the center block. Bottom: This architecture can predict a block given four of its neighbors (three blocks to the top and one to the left). We use this image prediction to compress an image. We first try to predict a block and then we store the residual which takes less space.}
\label{fig:splash}
\end{figure}

For image compression, we examine image blocks in a sequence. When each block is being compressed, we first try to predict the block's image according to its neighbouring blocks~(Figure~\ref{fig:splash}). Our prediction has a residual with respect to the original block. We store this residual which requires less space than the original block. We compress this residual using legacy JPEG techniques. We can trade off quality versus space using JPEG compression ratio.
Our image prediction is a deterministic process. Therefore, during decompression, we first try to predict a block's content and then add up the stored residual. After decompression, we perform artifact removal to further improve the quality of the restored image. With this technique we get a superior quality-space trade-off.

\begin{figure}[t]
\centering
    \includegraphics{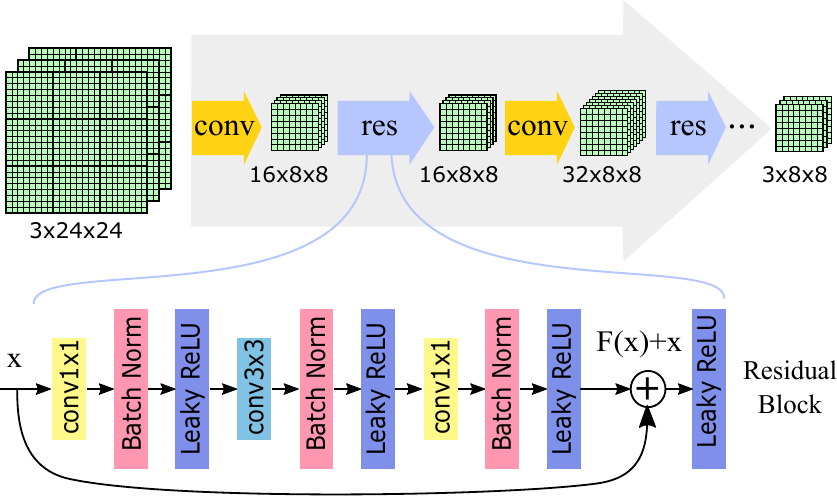}
    \caption{Our network architecture. Top: We input a $24 \times 24$ color image and output an $8 \times 8$ color image. Our network has a series of convolution and residual blocks. Bottom: Our residual block consists of several operations including convolution, batch normalization, and leaky ReLU activation function.}
\label{fig:arch}
\end{figure}

\subsection{Related Work}

JPEG \cite{Wallace:1991:JSP:103085.103089} compresses $8 \times 8$ blocks using quantized cosine coefficients. JPEG compression could lead to unwanted compression artifacts. Several techniques are developed to reduce artifacts and improve compression:

\begin{itemize}
\item {\bf Deep Learning}:
    Jain et al.~\cite{NaturalImageDenoisingwithConvolutionalNetworks} and Zhang et al.~\cite{DBLP:journals/corr/BeyondaGaussianDenoiser} trained a network to reduce Gaussian noise.
    This network does not need to know noise level.
    Dong et al.~\cite{DBLP:conf/iccv/CompressionArtifactsReductionbyaDeepConvolutionalNetwork} trained a network to reduce JPEG compression artifacts. 
    Ballé et al.~\cite{DBLP:journals/corr/EndtoendOptimized} uses a nonlinear analysis transformation, a uniform quantizer, and a nonlinear synthesis transformation. 
    Mao et al.~\cite{DBLP:conf/nips/ImageRestorationUsingVeryDeepConvolutional} developed an encoder-decoder network for denoising. This network uses skip connections and deconvolution layers. 
    Theis et al.~\cite{Lossyimagecompressionwithcompressiveautoencoders} presented an auto-encoder based compression technique. 
\item {\bf Residual-based Techniques}:
    Svoboda et al.~\cite{DBLP:journals/corr/CompressionArtifactsRemovalUsingConvolutionalNeuralNetworks} applied residual representation learning to define an easier task for network.
    Baig et al.~\cite{DBLP:journals/corr/LearningtoInpaintforImageCompression} use image inpainting before compression.
    Dong et al.~\cite{DBLP:conf/iccv/CompressionArtifactsReductionbyaDeepConvolutionalNetwork} reuse pre-trained models to speeds up learning.
\item {\bf Generative Techniques}: 
    Santurkar et al.~\cite{DBLP:journals/corr/GenerativeCompression} used Deep Generative models to reproduce image and video and remove artifacts. 
    A notable work in image generation is PixelCNN by Oord et al.~\cite{DBLP:journals/corr/PixelCNN}. Dahl et al.~\cite{DBLP:conf/iccv/PixelRecursiveSuperResolution} introduced a super-resolution technique based on PixelCNN. Our BlockCNN architecture is also inspired by PixelCNN.
\item {\bf Recurrent Neural Networks}:
    Toderici et al.~\cite{DBLP:conf/cvpr/FullResolutionImageCompressionwithRecurrentNeuralNetworks} presented a compression technique using an RNN-based encoder and decoder, binarizer, and a neural network for entropy coding. They also employ a new variation of Gated Recurrent Unit~\cite{DBLP:journals/corr/GRU}.
    Another work by Toderici et al.~\cite{DBLP:journals/corr/VariableRateImageCompressionwithRecurrentNeuralNetworks} proposed a variable-rate compression technique using convolutional and deconvolutional LSTM~\cite{lstm} network.
\end{itemize}

\begin{figure}[t]
\centering
    \includegraphics[width= 7.5cm]{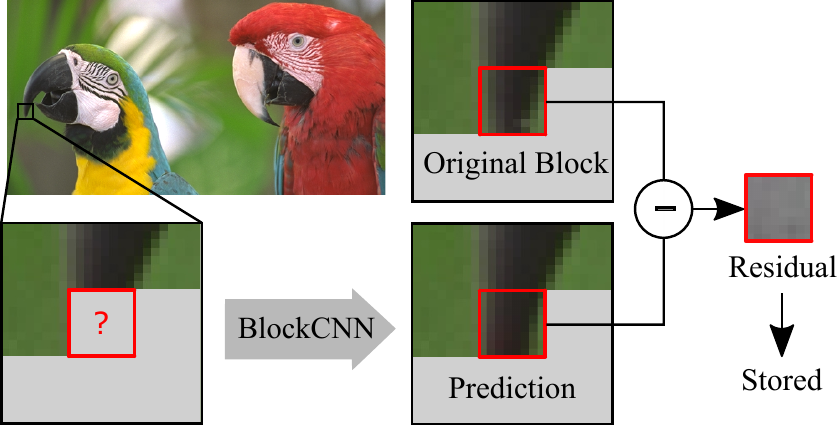}
    \caption{Our compression pipeline. We process image blocks in a sequence. For each block (highlighted with question mark), we first try to predict its intensities using the previous blocks. Then we compute the residual between our prediction and the original block. We store this residual and continue to the next block. During decompression we go through a similar sequential process. We first predict an image block using its previous blocks and then add up the residual.}
\label{fig:compression}
\end{figure}

\section{BlockCNN}

Similar to JPEG, we partition an image into $8 \times 8$ blocks and process each block separately. We use a convolutional neural network that inputs a block together with its adjacent blocks (a $24 \times 24$ image), and outputs the processed block in the center. We call this architecture BlockCNN. We use BlockCNN both for artifact removal and image compression. In the following subsections, we discuss the specifications in more detail.

\subsection{Deep Architecture}

BlockCNN consists of a number of convolution layers in the beginning followed by a number of residual blocks (resnet~\cite{DeepResNet}). A simplified schematic of the architecture is illustrated in Figure~\ref{fig:arch}. A residual block is formulated as:
\begin{equation}
    \label{resblock}
    G(x) = F(x) + x
\end{equation}
where $x$ shows the identity mapping and $F(x)$ is a feed-forward neural network trying to learn the residual (Figure~\ref{fig:arch}, bottom).  Residual blocks avoid over-fitting, vanishing gradient, and exploding gradient. Our experiments show that residual blocks are superior in terms of accuracy and rate of convergence. 


We use mean square error as loss function. For training, we use Adam~\cite{DBLP:journals/corr/Adam} with weight decay of $10^{-4}$ and learning rate of $10^{-3}$. We train our network for $120,000$ iterations. 


\begin{figure}[t]
\centering
    \includegraphics{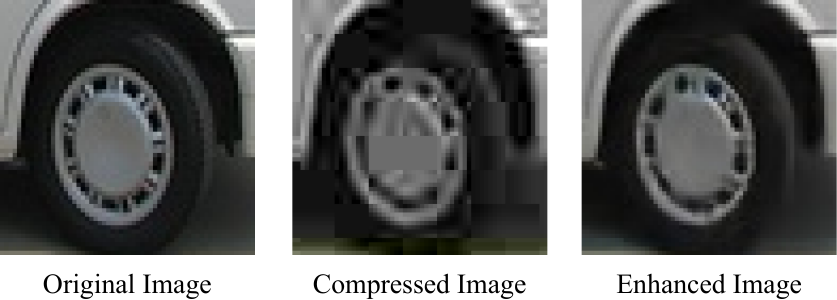}
    \caption{
    Left: Original image before compression. Center: JPEG compressed Image with block-wise artifact. Right: Enhanced Image using our technique. Note that our result has improved blocking and ringing artifacts.}
\label{fig:artifact}
\end{figure}

\begin{figure}[t]
\centering
    \includegraphics{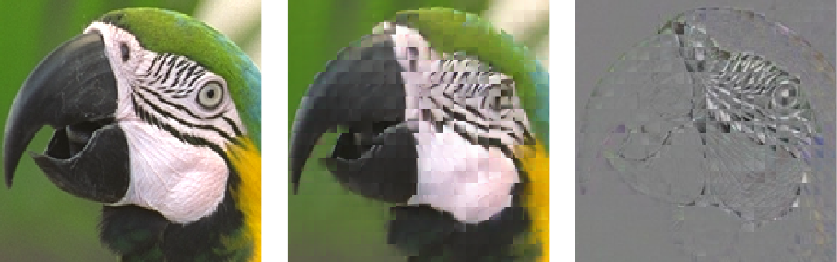}
    \caption{
    Left: Original Image fed to BlockCNN for compression. Center: BlockCNN prediction. Each block in this image shows the best prediction using previously seen blocks. Right: The difference between the original image and our prediction~(Residual). We store residual instead of the original image. 
    }
\label{fig:PixelCNN}
\end{figure}

\subsection{BlockCNN for Artifact Removal}
\label{ssec:artifact}

For artifact removal we train BlockCNN with JPEG compressed blocks as input and an uncompressed block as target~(Figure~\ref{fig:splash}). 
This network has three characteristics that make it successful.

\begin{itemize}
\item {\bf Residual}: Since compression artifact is naturally a residual, predicting artifact as residual is easier than encoding and decoding a block. It leads to faster and improved convergence behavior.
\item {\bf Context}: BlockCNN input adjacent cells so it can use context to improve its estimate of residual. Of course, larger context can improve performance but we limit context to one block for better illustration.
\item {\bf block structure}: The statistics of artifact depend on where a pixel is placed within a block. BlockCNN takes advantage of the prior of pixels' location within a block.
\end{itemize}

\subsection{BlockCNN for Image Compression}
\label{sec:blockcnn}

The idea behind our technique is the following: Anything that can be predicted does not need to be stored.

We traverse image blocks row wise. Given each block, we first try to predict its intensities given previously seen blocks. Then we compute the residual between our prediction and the actual block and store this residual~(Figure~\ref{fig:compression}). When storing a block's residual we compress it using JPEG. JPEG has two major benefits: 1- It is simple, fast and readily available; 2- We can reuse JPEG's quality factor to trade off size with quality.

During decompression we follow the same deterministic process. For each block, we try to predict its intensities and then add up the stored residual. Note that for predicting a block's image, we  use the {\it compressed} version of the previous blocks, because otherwise compression noise propagates and accumulates throughout image. For this prediction process we reuse BlockCNN architecture but with different training examples and thus different weights. The major difference in training examples is that only four out of nine blocks are given as input.

\begin{figure}[t]
\centering
    \includegraphics[width = 8.2 cm]{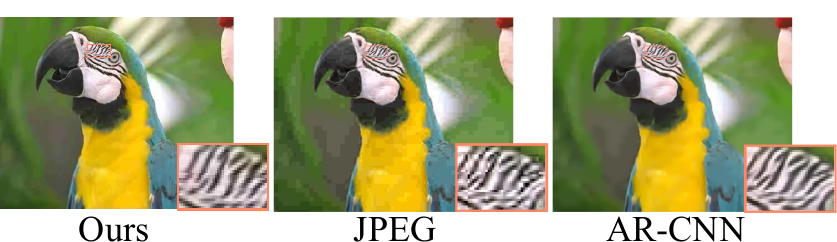}
    \caption{
    Qualitative comparison between artifact removal algorithms.}
\label{fig:ArtifactRemovalResults}
\end{figure}

\begin{figure*}[t]
\centering
    \includegraphics[width = 16 cm]{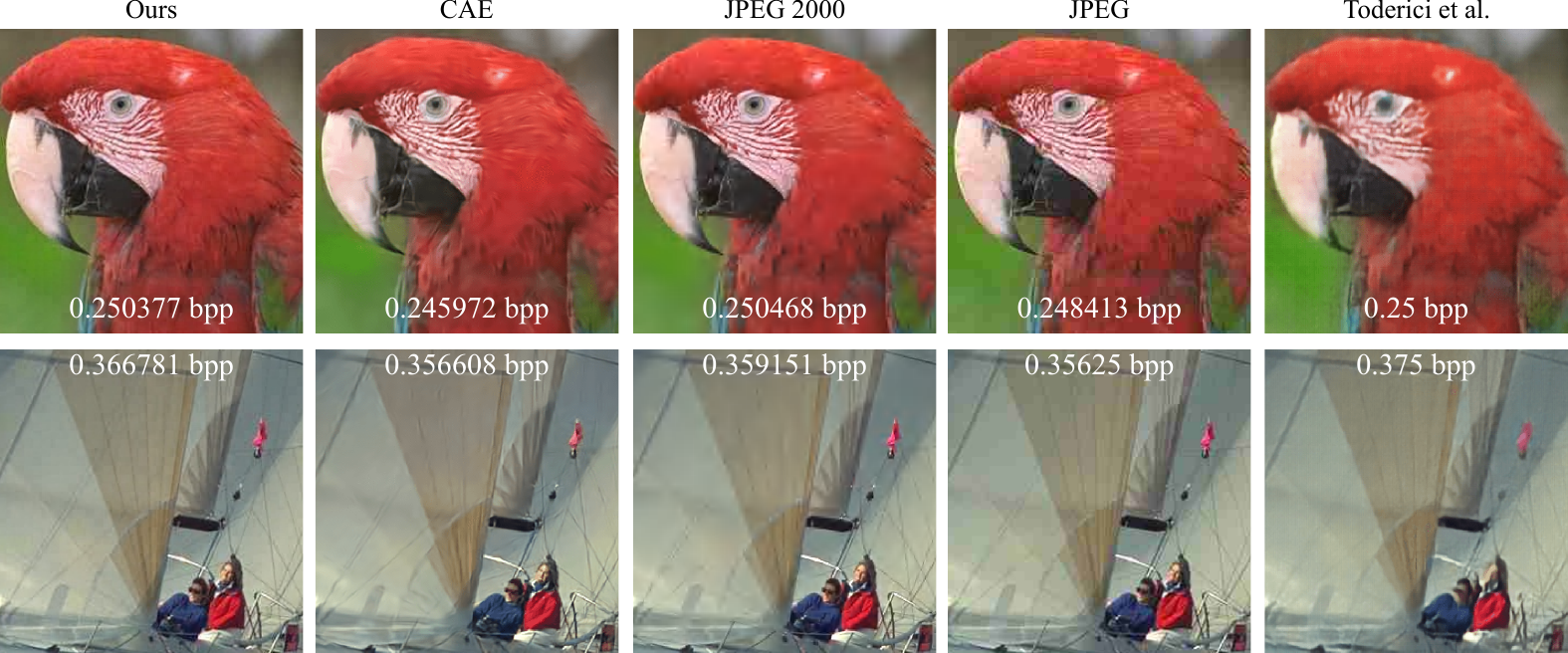}
    \caption{
    Qualitative comparison between different compression algorithms at low bit rates. Note that our technique performs a better job at preserving details.
    }
\label{fig:CompressionResults}
\end{figure*}

\section{Experiments and Results}

We used PASCAL VOC~2007~\cite{Pascal} for training. Since JPEG compresses images in Lab colorspace, it produces separate intensity and color artifacts. Therefore, we perform training in Lab colorspace to improve our performance for ringing, blocking, and chromatic distortions. 

For optimization process we use Adam~\cite{DBLP:journals/corr/Adam} with weight decay of $10^{-4}$ and a learning rate of $10^{-3}$. This network is trained for $120,000$ iterations. 

\subsection{Artifact Removal Results}

Artifact removal techniques in the literature are usually benchmarked using 
LIVE~\cite{live1} dataset. Peak signal-to-noise ratio~(P-SNR) and structural similarity index measurement (SSIM)~\cite{SSIM} are regularly used as evaluation criteria. We follow this convention and compare our results with AR-CNN~\cite{DBLP:conf/iccv/CompressionArtifactsReductionbyaDeepConvolutionalNetwork}~(Figure~\ref{fig:ArtifactRemovalResults}).
We trained a BlockCNN with $9$ residual blocks. We use approximately 6 Million pairs of input and target for training.

\begin{figure}[t]
\centering
    \includegraphics[width = 8.2 cm, height = 6.2cm]{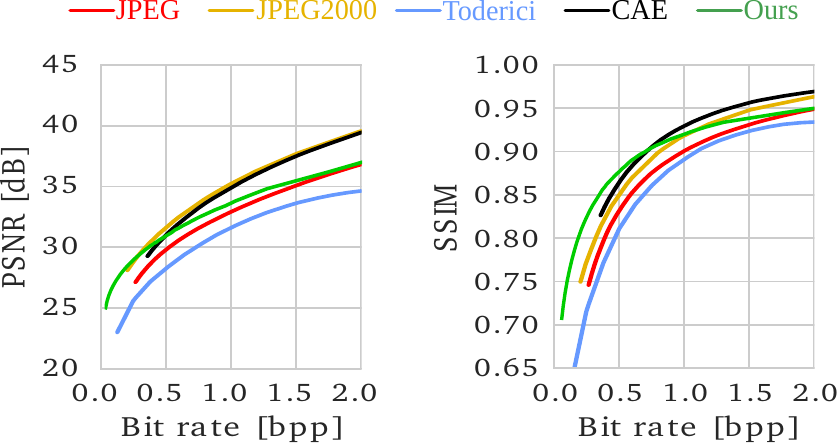}
    \caption{
    Comparison of compression techniques using PSNR and SSIM. Our technique outperforms baseline at low bit rates. This is because highly compressed blocks only store low frequency information that is predictable. At higher bit rates it relies more heavily on JPEG compression routines. Therefore our improvement diminishes at higher bit rates.
    }
\label{fig:CompressionBenchmark}
\end{figure}

\begin{figure}[h]
\centering
    \includegraphics[width = 8.2cm]{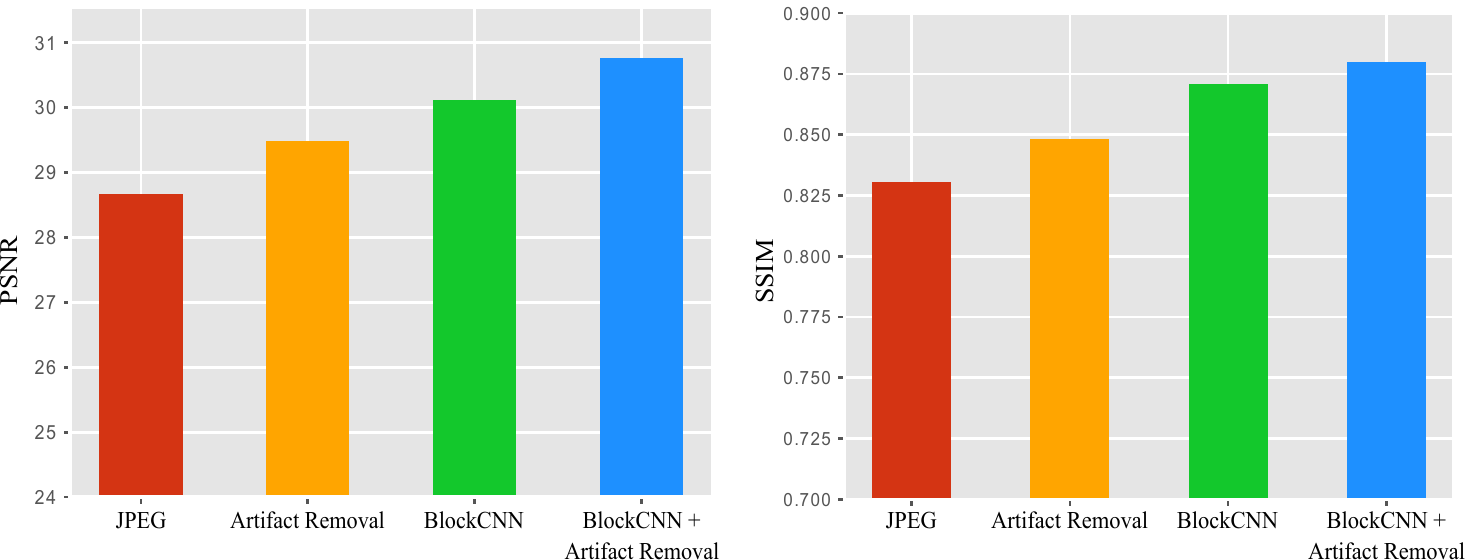}
    \caption{
    We can independently switch on or off our compression and artifact removal steps. This figure compares the four combinations using P-SNR and SSIM. Note that the improvements from the two steps are independent. This figure is generated using kodak dataset with $bpp = 0.45$.
    }
\label{fig:barplot}
\end{figure}

\subsection{Image Compression Results}

Kodak~\footnote{http://r0k.us/graphics/kodak/} dataset is commonly used as a benchmark for Image Compression. We evaluate our method using peak signal-to-noise ratio~(P-SNR) and structural similarity index measurement~(SSIM). After image decompression, we perform artifact removal to improve quality. We compare our results with Toderici et al.~\cite{DBLP:conf/cvpr/FullResolutionImageCompressionwithRecurrentNeuralNetworks} and CAE~\cite{Lossyimagecompressionwithcompressiveautoencoders}~(Figure~\ref{fig:CompressionBenchmark}).

Our image compression algorithm tries to minimize storage by predicting the content of blocks. Predicting high-frequency details is difficult and discouraged by most loss functions. As a result, our algorithm performs a better job at low compression factors that do not predict high frequency details.

\subsection{Image Compression and Artifact Removal}

We can switch on and off image compression and artifact removal steps. Therefore, we have four combinations of techniques. (Figure~\ref{fig:barplot})
\begin{enumerate}
\item Original JPEG results without encorporating BlockCNN or artifact removal process.
\item Using artifact removal without BlockCNN. This method inputs a previously saved JPEG image and tries to enhance it~(Section~\ref{ssec:artifact}).
\item Using BlockCNN without artifact removal. We compress the input image according to Section~\ref{sec:blockcnn}. However, we do not perform artifact removal. This process leads to better P-SNR and SSIM given the same bpp.
\item Using both BlockCNN and artifact removal. The improvements from BlockCNN compression and artifact removal are independent. Therefore, by using both artifact removal and BlockCNN we achieve better results.
\end{enumerate}

\section{Discussion}

We presented BlockCNN, a deep architecture that can perform artifact removal and image compression. Our technique respects JPEG compression conventions and acts on $8 \times 8$ blocks. The idea behind our image compression technique is that before compressing each block, we try to predict as much as possible from previously seen blocks. Then, we only store a residual that takes less space. Our technique reuses JPEG compression routines to compress the residual. Our technique is simple but effective and it beats baselines for high compression ratios.

{\small
\bibliographystyle{ieee}
\bibliography{egbib}
}

\end{document}